\pdfoutput=1

\documentclass[11pt]{article}

\usepackage[]{acl}

\usepackage{times}
\usepackage{latexsym}
\usepackage{graphicx}
\usepackage{tabularx}
\usepackage{colortbl}
\usepackage[T1]{fontenc}

\usepackage[utf8]{inputenc}

\usepackage{microtype}
\usepackage{paralist}
\usepackage{diagbox}
\usepackage{multirow}
\usepackage{comment}
\usepackage{booktabs}
\usepackage{relsize}

\newcommand{\email}[1]{{\texttt{\smaller #1}}}

\title{Improving negation detection with negation-focused pre-training}

\author{Hung Thinh Truong$^1$ Timothy Baldwin$^{1,3}$ Trevor Cohn$^1$ Karin Verspoor$^2$\\
$^1$The University of Melbourne, $^2$RMIT University, $^3$MBZUAI\\
\email{hungthinht@student.unimelb.edu.au},  \email{tb@ldwin.net}, \\ \email{trevor.cohn@unimelb.edu.au}, \email{karin.verspoor@rmit.edu.au}}

\begin{document}

\maketitle
\begin{abstract}

Negation is a common linguistic feature that is crucial in many language understanding tasks, yet it remains a hard problem due to diversity in its expression in different types of text. 
Recent work has shown that state-of-the-art NLP models underperform on samples containing negation in various tasks, and that negation detection models do not transfer well across domains.
We propose a new negation-focused pre-training strategy, involving targeted data augmentation and negation masking, to better incorporate negation information into language models.
Extensive experiments on common benchmarks show that our proposed approach improves negation detection performance and generalizability over the strong baseline NegBERT \cite{khandelwal2020negbert}.

\end{abstract}

\section{Introduction}
\label{sec:intro}
Negation is an important linguistic phenomenon that appears commonly in natural language but is 
underrepresented in common NLP benchmarks \cite{hossain2020analysis}.
Furthermore, the Checklist benchmark \cite{ribeiro2020beyond} shows that most sentiment analyzers and machine comprehension models struggle with samples containing negation.
Negation is even more important in biomedical domain text, where 
patients are carefully defined as having/not having specific characteristics.
Even within the biomedical domain, there are many types of text such as clinical notes, lab reports, or research publications, each with particular characteristics in relation to the use of negation.
A recent study on English texts found that negation detection models do not transfer well across domains, due to variations in expression of negation  \citep{khandelwal2020negbert}.
It remains a challenge to solve negation in general, even with state-of-the-art NLP models. 

Negation detection is typically defined as consisting of the two sub-tasks of: (1) cue detection, detecting the cue phrase that triggers the negation; and (2) scope resolution, determining the affected spans that are negated. 
There are three primary datasets that have been used to evaluate negation:
(1) the \textit{BioScope} corpus \cite{vincze2008bioscope} includes full
papers and abstracts of biological papers; (2) the \textit{SFU} corpus
\cite{konstantinova-etal-2012-review} is a collection of product reviews; and (3) the \textit{Sherlock} dataset \cite{morante2012sem} consists of short literary works. 
There are differences in annotation schemes across the datasets, such as whether or not the cues are included inside scope annotation,
and sub-optimal cross-dataset results have been observed, providing clear indications that the datasets are highly divergent in language use and negation types. 

In this work, we aim to extend the transfer learning capability of NegBERT \citep{khandelwal2020negbert}  through additional pre-training with task-related augmented training data, 
and a new masking objective. 
Our contributions are:
\begin{compactitem}
    \item We introduce an approach to augmenting data to emphasize negation in pre-training.
    \item We propose a novel extension to the standard random masked language model objective in pre-training to explicitly mask negation cues, to make the models more robust to negation.
    \item We conduct extensive experiments on different benchmarks to
      evaluate cross-domain performance of large pre-trained language
      models as well as the effectiveness of the proposed pre-training
      strategies; code is available at \url{https://github.com/joey234/negation-focused-pretraining}.
\end{compactitem}

\section{Related work}
To date, negation detection has been heavily reliant on rule-based systems.
\citet{chapman2001simple} proposed a simple system, NegEx, based on regular expressions to detect negation cues in a sentence given a concept of interest (the scope).
NegEx remains the most popular approach to negation detection, especially in the clinical domain 
to determine the polarity of clinical concepts (e.g., as sourced from MetaMap \citep{aronson2010metamap}).
Further research has extended NegEx with syntactic information \citep{mehrabi2015deepen,peng2018negbio}, and
shown that rule-based systems can achieve relatively good performance for detecting negation, especially in the biomedical domain, but do not generalize well to other domains or datasets. 

To approach negation cue and negation scope detection with supervised machine learning,
two classification tasks are defined: (1) finding negation tokens, and (2) classifying tokens as the first or last (or neither) token within the scope of negation. 
Most work follows a common scheme in extracting various features from the sentence, and using a classifier to classify each token as the beginning, inside, or outside of a negation cue or scope span \citep{morante2009metalearning,ou2015automatic,cruz2016machine}. 
Recently, research has shifted to applying deep learning methods to the task. Most approaches make use of RNN-based architectures to encode the input sentences, combined with a softmax layer for classification \citep{lazib2019negation, chen2019attention}. Despite the high performance on common benchmarks, results are biased by the fact that negation scope is often delimited by punctuation and other dataset artefacts \citep{fancellu2017detecting}. As such, they are potentially only learning domain-specific surface features rather than capturing the true semantics of negation.
NegBERT applies a large pre-trained language model to the problem of negation detection,  outperforming previous deep learning methods on negation detection, with especially high gains on scope resolution. 

\section{Method}
Our proposed pre-training strategy consists of two main components: (1) negation-focused data collection in which we first collect relevant data that contains negation; and 
(2) negation-focused pre-training that makes use of the negation-focused data to emphasize negation instances, and adopts a novel negation-specific masking strategy.

\subsection{Negation-focused data collection}
We aim to construct a dataset that is enriched with negation information, to
support negation-sensitized pre-training of large language models. 
To obtain sentences with negations, we extend the NegEx 
lexicon with additional negation cues obtained from biomedical texts \cite{morante2010descriptive}, and apply it to sentences extracted from a corpus using the SpaCy English sentence tokenizer, keeping only those sentences with at least one identified negation cue.

For the biomedical domain, we use texts in the TREC-CDS 2021 snapshot\footnote{{\footnotesize \url{http://www.trec-cds.org/2021.html}}} of the clinical trials registry.\footnote{\footnotesize\url{http://clinicalTrials.gov}}
Clinical trials are documents describing the protocols and relevant patient characteristics of a clinical research study.
Descriptions of clinical trials can be quite long, but a core aspect of the trial description is the patient inclusion/exclusion criteria, specifying what types of characteristics or conditions a patient must have/not have in order to be suitable for the trial.
The reasons for choosing this data are that: (1) it is in-domain for the biomedical domain; (2) the texts are well-formed sentences with proper grammatical structure; and (3) the texts contain many negations, especially in the inclusion/exclusion criteria sections.
For the general domain, we apply this approach to \textit{wikitext} \cite{merity2016pointer}, a set of verified 
articles in Wikipedia. 
We sample the data equally from these two sets, obtaining $1,381,948$ negation sentences.

\subsection{Negation-focused pre-training}
Adaptive pre-training on target domain data has been shown to be an effective strategy for domain adaptation \cite{gururangan2020don}.
We therefore hypothesize that pre-training language models on text with negations will help the model incorporate information about negation, and learn better representations for sentences containing negation.
Using the negation-focused data, we first apply the standard random word masking strategy
\cite{devlin-etal-2019-bert} and train the model with the masked language model objective. 

As part of the collection of the negation-focused data, we obtain predictions of negation cues in all the sentences, which can be explicitly incorporated to make the model more robust to negation. 
Inspired by work on entity and span masking \cite{joshi2020spanbert,yamada2020luke}, we explore explicitly incorporating information about negation cues into the model by masking  these cues, and targeting prediction of the masked cue in the pre-training stage. 
Below is an example of how a sentence is tokenized under our masking scheme:\\[0.5ex]
\textit{No serious complications such as hypertension, diabetes.} $\Rightarrow$
\textit{\textbf{[CUE]} serious complications such as \textbf{[MASK]}, diabetes.}

A new type of token \textit{\textbf{[CUE]}} is introduced under this masking scheme, and the model needs to reconstruct the original sentence by predicting both the \textit{\textbf{[CUE]}} token to be \textit{No}, and the randomly-masked token \textit{\textbf{[MASK]}} to be \textit{hypertension}.
By always masking negation cues in all the sentences, we force the model to focus more on this type of token, and thus, aim to learn better embeddings incorporating information of how a negation cue is represented in the context of the sentence.
Moreover, by using a different token to mask negation cues, we ensure that the model learns to distinguish between different types of tokens. 
In this work, we replace the BERT encoder of NegBERT with RoBERTa \cite{liu2019roberta} and apply whole-word masking, meaning that all the sub-word tokens that constitute a word will be masked.

\section{Experiments}
\subsection{Experimental settings}

Following the experimental settings in NegBERT, 
we use the three standard benchmarks for negation cue detection and scope resolution tasks, i.e.\ BioScope \cite{vincze2008bioscope} (separated into two subsets, sourced from abstracts and full-text papers, resp.), the SFU product reviews dataset \cite{konstantinova-etal-2012-review}, and the Sherlock dataset \cite{morante2012sem}.
In addition, we use the negation-annotated subset of VetCompass UK\footnote{\url{https://www.rvc.ac.uk/VetCOMPASS}} 
\citep{cheng2017automatic}, 
consisting of clinical notes in the veterinary domain, which are very informal compared to BioScope. It also contains abbreviations and shortening of terms, as well as certain unique negation cues. 
To investigate cross-domain performance, we perform cue detection and scope resolution for all 4 datasets, based on training on one dataset and evaluating on all datasets.
Detailed statistics of these datasets are presented in Table~\ref{tab:statistics}.
Note that we do not experiment with the setting of training with all the combined training data from the corpora as it has been pointed out by previous work  that doing so hurts the performance of the models \citep{jimenez-zafra-etal-2020-corpora,barnes2021improving} due to differences in annotation schemes between the corpora introducing noise during training.
Re-annotating all the datasets using a common annotation scheme would be a potential solution here, which we leave for future work.

We formulate the two tasks as sequence labeling problems, where each token is tagged with a corresponding label.
For cue detection, we use the annotation scheme \textit{\{0: Affix, 1: Normal Cue, 2: Part of multiword cue, 3: Not part of cue\}}.
For scope resolution, we use gold cue information and two labels \textit{\{0: Outside negation scope, 1: Part of negation scope\}}.
We adopt the same hyperparameters as NegBERT. 
Following the standard evaluation scheme in previous negation detection works, all systems are evaluated using token-level $F_{1}$-score, based on whether it is inside or outside of any negation cue or scope.
Methods evaluated include: (1) \textbf{NegBERT}; (2) \textbf{AugNB} = NegBERT plus pre-training on negation-focused data; and (3) \textbf{CueNB} = NegBERT plus pre-training on negation-focused data and the negation cue masking objective.
Note that for NegBERT, we also replace the BERT encoder with RoBERTa to ensure results are comparable between the models.

\begin{table}[!t]
    \centering
        \begin{tabular}{p{1.75cm} p{1.5cm} p{1.5cm} p{1.25cm}}
    \toprule
        \textbf{Dataset} & \textbf{\#sentences} & \textbf{\#negations} & \textbf{\#unique cues}  \\
        \midrule
        BioScope-Abstract & 11871 & 1719  & 28 \\
        BioScope-FullPaper & 2670 & 376 & 18 \\ 
        SFU & 17263 & 3527 & 53 \\ 
        Sherlock & 5520 & 1421 & 30 \\
        VetCompass & 6582 & 724 & 26 \\
        \bottomrule    
        \end{tabular}
    \caption{Dataset statistics}
    \label{tab:statistics}

\end{table}




\subsection{Main results}
Tables~\ref{tab:cue_res} and \ref{tab:scope_res} report the performance of negation cue detection and negation scope resolution,  respectively. 
Results reported are the average of 5 runs with different random seeds.  NegBERT results are produced using the official implementation.\footnote{\footnotesize\url{https://github.com/adityak6798/Transformers-For-Negation-and-Speculation}}
To provide a more general view, we summarize the results in Table~\ref{tab:aggregate_results}.
In general, we observe gains in both the same-dataset setting (training and test set belongs to one corpus) and cross-dataset setting (training one one training set and testing on all others test sets) for both of the proposed models, with CueNB achieving the largest gains.

\begin{table*}[!t]
    \centering
    \setlength{\tabcolsep}{0.3em}
    \resizebox{16cm}{!}{
    

    \begin{tabular}{c| rrr | rrr | rrr | rrr | rrr }
        
        \multirow{2}{*}{\backslashbox{Evaluation set}{Training set}}  & \multicolumn{3}{c|}{BioScope-Abstract} & \multicolumn{3}{c|}{BioScope-FullPaper} & \multicolumn{3}{c|}{SFU} & \multicolumn{3}{c|}{Sherlock} & \multicolumn{3}{c}{VetCompass}  \\
        \cmidrule{2-16}
         & NegBERT & AugNB & CueNB & NegBERT & AugNB & CueNB & NegBERT & AugNB & CueNB & NegBERT & AugNB & CueNB & NegBERT & AugNB & CueNB \\
        \midrule
        BioScope-Abstract & \cellcolor{gray!25} 95.29 & \cellcolor{gray!25} $+$0.78 & \cellcolor{green!25} $+$1.80 & 92.58 & $-$0.27 & $-$0.25 & 83.86 & $+$0.23 & $+$0.47 & 70.67 & $+$4.89 & $+$6.62 & 75.49 & $+$1.56 & $+$2.73 \\
        BioScope-FullPaper & 91.44 & $+$0.68 & \cellcolor{green!25} $+$0.89 & \cellcolor{gray!25} 90.23 & \cellcolor{gray!25} $+$0.04 & \cellcolor{gray!25} $+$1.44 & 79.68 & $+$0.11 & $+$0.63 & 66.45 & $+$1.07 & $+$2.48 & 71.06  & $+$1.17  & $+$2.98  \\
        SFU & 38.70 & $+$3.55 & $+$4.78  & 57.99 & $+$3.72 & $+$4.43 & \cellcolor{gray!25} 87.20 & \cellcolor{gray!25} $+$0.38 & \cellcolor{green!25} $+$0.79 & 44.03 & $+$11.37 & $+$13.93 & 58.66 & $+$0.67 & $+$2.44  \\
        Sherlock & 70.43 & $+$2.86 & $+$3.05 & 69.63 & $+$4.54 & $+$6.48 & 70.14 & $+$1.26 & $+$1.79 & \cellcolor{green!25} 92.28 & \cellcolor{gray!25} \textcolor{red}{$-$0.51} & \cellcolor{gray!25} \textcolor{red}{$-$1.11} & 64.45 & $+$3.58 & $+$3.99 \\
        VetCompass & 70.58 & $+$0.37 & $+$1.91 & 69.75 & $+$0.36 & $+$2.39 & 75.18 & $+$2.19 & $+$3.42 & 71.34 & $+$0.33 & $+$1.07 & \cellcolor{gray!25} 87.77 & \cellcolor{gray!25} $+$1.11 & \cellcolor{green!25} $+$3.77 \\
    \end{tabular}
    }
    \caption{Cue detection results. Gray cells denote the same-dataset setting, and green cells indicate the highest score for each evaluation dataset. Results of AugNB and CueNB are relative changes compared to NegBERT}
    \label{tab:cue_res}
\end{table*}

\begin{table*}[!t]
    \centering
    \setlength{\tabcolsep}{0.3em}
    \resizebox{16cm}{!}{

    \begin{tabular}{c| rrr | rrr | rrr | rrr | rrr }
        
        \multirow{2}{*}{\backslashbox{Evaluation set}{Training set}}  & \multicolumn{3}{c|}{BioScope-Abstract} & \multicolumn{3}{c|}{BioScope-FullPaper} & \multicolumn{3}{c|}{SFU} & \multicolumn{3}{c|}{Sherlock} & \multicolumn{3}{c}{VetCompass}  \\
        \cmidrule{2-16}

         & NegBERT & AugNB & CueNB & NegBERT & AugNB & CueNB & NegBERT & AugNB & CueNB & NegBERT & AugNB & CueNB & NegBERT & AugNB & CueNB \\
        \hline
        BioScope-Abstract & \cellcolor{gray!25} 94.23 & \cellcolor{gray!25} $+$0.84  & \cellcolor{green!25} $+$1.58 & 90.89 & $+$0.70  & $+$0.74  & 84.41 & $+$0.15  & $+$0.43 & 78.80 & $+$0.63  & $+$1.66 & 69.14 & $+$1.57  & $+$2.82  \\
        BioScope-FullPaper & 91.63 & $+$1.14  & \cellcolor{green!25} $+$1.83 & \cellcolor{gray!25} 88.42 & \cellcolor{gray!25} $+$1.80 & \cellcolor{gray!25} $+$4.14 & 79.90 & $+$0.49 & $+$0.83 & 79.42 & $+$0.21 & $+$1.28 & 64.45 & $+$1.79  & $+$2.27 \\
        SFU & 85.28 & $+$0.78 & $+$1.03 & 84.57 & $+$0.71 & $+$1.05 & \cellcolor{gray!25} 90.44 & \cellcolor{gray!25} $+$0.27 & \cellcolor{green!25} $+$0.59 & 74.61 & $+$1.88 & $+$3.28 & 63.32  & $+$3.25  & $+$3.59 \\
        Sherlock & 72.60 & $+$0.43 & $+$2.17 & 70.10 & $+$2.24 & $+$3.04 & 73.68 & $+$0.11 & $+$0.87 & \cellcolor{green!25} 91.51 & \cellcolor{gray!25} \textcolor{red}{$-$1.20}  & \cellcolor{gray!25} \textcolor{red}{$-$0.27} & 61.49 & \textcolor{red}{$-$0.03 } & $+$1.64  \\
        VetCompass & 61.36 & $+$0.86 & $+$2.00 & 60.27  & $+$1.06 & $+$1.39 & 62.62 & $+$0.32  & $+$1.40  & 59.62 & $+$0.61  & $+$1.05   & \cellcolor{gray!25} 88.18 & \cellcolor{gray!25} $+$1.23 & \cellcolor{green!25} $+$2.06 \\
    \end{tabular}
    }
    \caption{Scope resolution results. Gray calls denote the same-dataset setting, and green cells indicate the highest score for each evaluation dataset. Results of AugNB and CueNB are relative changes compared to NegBERT}
    \label{tab:scope_res}
\end{table*}

\begin{table*}[!t]
    \centering
    \begin{tabular}{c ccc c ccc}
      \toprule
      \multirow{2}{*}{Task} & \multicolumn{3}{c}{Same-dataset results} && \multicolumn{3}{c}{Cross-dataset results}  \\
      \cmidrule{2-4}
      \cmidrule{6-8}
        & NegBERT &  AugNB &  CueNB && NegBERT &  AugNB &  CueNB \\  
        \midrule
        Cue Detection & 90.55 & $+$0.36 & $+$1.34 && 69.61 & $+$2.21 & $+$ 3.31 \\
        Scope Resolution & 90.56 & $+$ 0.59 & $+$1.62 && 73.41 & $+$0.95 & $+$ 1.72 \\
      \bottomrule
    \end{tabular}
    \caption{Aggregated results}
    \label{tab:aggregate_results}
\end{table*}

We observe similar trends across all datasets for both cue detection and scope resolution.
Regarding the in-dataset setting, AugNB outperforms the baseline NegBERT on all datasets except for Sherlock.
Gains are more noticeable over the biomedical datasets (BioScope, VetCompass).
For Sherlock, however, we observe a slight degradation in performance with the proposed pre-training scheme.
This is likely due to the fact that Sherlock has major differences in annotation scheme compared to other corpora, specifically including scopes to the left of cues, while in BioScope and SFU, the scope is usually annotated only to the right of  cues. Also, the cue itself is not considered to be part of the scope in Sherlock or SFU, unlike in BioScope.

In the cross-dataset setting, we record gains across all benchmarks. The largest cross-dataset improvements over NegBERT are for SFU, perhaps due to SFU being the largest dataset in size, containing a relatively large number of unique cues.
CueNB further improves the performance of AugNB, confirming our hypothesis that explicitly masking the cue helps the model learn better representations for negation cues and thus, better distinguish between cues and normal words.
These results show that our negation-focused pre-training strategy is effective for improving the transfer learning performance of pre-trained language models on the negation detection task.

\subsection{Discussion}

We conducted an error analysis on the VetCompass validation set to see what qualitative improvement CueNB makes over NegBERT.
For cue detection, there are two main types of errors that CueNB helps alleviate.
First, CueNB can detect more unique cues such as \textit{negative}, \textit{won't}, 
 and also multiword cues like \textit{no longer}.
Second, CueNB is able to recognize cases when the negations are actually just speculative. For example, in the sentence \textit{O reports has smelled for past week, \textbf{not sure} if anal glands \ldots}, the word \textit{not} is  part of the speculation phrase \textit{not sure}, indicating that this is not truly a negation phrase but rather expresses uncertainty.
For scope resolution, CueNB mostly helps in recognizing the correct scope boundary. 
One common case is when the cue relates to multiple spans in a sentence.
In the sentence \textit{Examination: QAR, thorac ausc and abdo palp NAD},\footnote{\textit{NAD} is the negation cue \textit{no abnormality detected}, \textit{QAR}, \textit{thorac ausc} (thoracic ausculation), \textit{abdo palp} (abdonimal palpation), are different types of physical examinations.}
NegBERT only recognizes the nearest span \textit{abdo palp NAD} to be the scope, whereas CueNB recognizes the full correct span \textit{thorac ausc and abdo palp NAD}.
It also helps in cases where there are multiple separate negations in the same sentence. 
For instance, in the sentence \textit{No V$+$ or no D$+$.}, the phrases \textit{No V$+$} and \textit{no D$+$} are two independent negation scope spans, while NegBERT would recognize the whole sentence as a single span.
Another interesting case is when there are exceptions in the sentence, e.g.\ the \textit{No \ldots other than \ldots} construction. 
For \textit{No probs detected other than the skin lesions}, CueNB is able to recognize the correct scope \textit{No probs detected} while NegBERT considers the whole sentence to be the scope.

We also conduct an ablation study to understand the impact of each component of the proposed pre-training strategy. Table~\ref{tab:ablation} presents the results of different variations of the proposed pre-training scheme on the BioScope-Abstract validation split.
We consider two variations, pre-training with: (1) only the negation-focused data (equivalent to the AugNB model); and (2) only the cue masking objective.
To model the latter variation, we explicitly mask the cue in the BioScope training set, then pre-train on this training set. 
From the results, we see that both strategies help improve the baseline NegBERT on cue detection and scope resolution, with explicitly masking the cues being the most important.
Combining both strategies (CueNB) further improves the overall results.


\begin{table}[!t]
    \centering
    \begin{tabularx}{\columnwidth}{Xp{1cm}p{1cm}}
      \toprule
        Model & Cue & Scope  \\
        \midrule
        NegBERT & 94.46 & 95.34 \\
        $+$ negation-focused data & 95.36 & 95.94  \\
        $+$ explicit cue masking & 95.58  & 96.03 \\
        \midrule
        CueNB & \textbf{95.87} & \textbf{96.76}
    \end{tabularx}
    \caption{Ablation study on BioScope validation set.}
    \label{tab:ablation}
\end{table}

%


\section{Conclusion}
In this work, we propose a new negation-focused pre-training strategy to explicitly incorporate negation information into pre-trained language models.
Empirical results on common benchmarks show that the proposed strategy helps improve the performance of pre-trained language models on the negation detection task when evaluating on the same source dataset, as well as their transferability to target data in different domains.
Despite the gains over previous methods, the sub-optimal results on some benchmarks show that negation remains a big challenge in NLP.

\section*{Acknowledgements}
The authors would like to thank the anonymous reviewers for their constructive reviews.
This research was undertaken using the LIEF HPC-GPGPU Facility hosted at the University of Melbourne. This Facility was established with the assistance of LIEF Grant LE170100200.
This research was conducted by the Australian Research Council Training Centre in Cognitive Computing for Medical Technologies (project number ICI70200030) and funded by the Australian Government.

\bibliography{custom}

\begin{thebibliography}{26}
\expandafter\ifx\csname natexlab\endcsname\relax\def\natexlab#1{#1}\fi

\bibitem[{Aronson and Lang(2010)}]{aronson2010metamap}
Alan~R Aronson and Fran{\c{c}}ois-Michel Lang. 2010.
\newblock An overview of {MetaMap}: {H}istorical perspective and recent
  advances.
\newblock \emph{Journal of the American Medical Informatics Association},
  17(3):229--236.

\bibitem[{Barnes et~al.(2021)Barnes, Velldal, and
  {\O}vrelid}]{barnes2021improving}
Jeremy Barnes, Erik Velldal, and Lilja {\O}vrelid. 2021.
\newblock Improving sentiment analysis with multi-task learning of negation.
\newblock \emph{Natural Language Engineering}, 27(2):249--269.

\bibitem[{Chapman et~al.(2001)Chapman, Bridewell, Hanbury, Cooper, and
  Buchanan}]{chapman2001simple}
Wendy~W Chapman, Will Bridewell, Paul Hanbury, Gregory~F Cooper, and Bruce~G
  Buchanan. 2001.
\newblock A simple algorithm for identifying negated findings and diseases in
  discharge summaries.
\newblock \emph{Journal of biomedical informatics}, 34(5):301--310.

\bibitem[{Chen(2019)}]{chen2019attention}
Long Chen. 2019.
\newblock Attention-based deep learning system for negation and assertion
  detection in clinical notes.
\newblock \emph{International Journal of Artificial Intelligence and
  Applications (IJAIA)}, 10(1).

\bibitem[{Cheng et~al.(2017)Cheng, Baldwin, and Verspoor}]{cheng2017automatic}
Katherine Cheng, Timothy Baldwin, and Karin Verspoor. 2017.
\newblock Automatic negation and speculation detection in veterinary clinical
  text.
\newblock In \emph{Proceedings of the Australasian Language Technology
  Association Workshop 2017}, pages 70--78.

\bibitem[{Cruz et~al.(2016)Cruz, Taboada, and Mitkov}]{cruz2016machine}
Noa~P Cruz, Maite Taboada, and Ruslan Mitkov. 2016.
\newblock A machine-learning approach to negation and speculation detection for
  sentiment analysis.
\newblock \emph{Journal of the Association for Information Science and
  Technology}, 67(9):2118--2136.

\bibitem[{Devlin et~al.(2019)Devlin, Chang, Lee, and
  Toutanova}]{devlin-etal-2019-bert}
Jacob Devlin, Ming-Wei Chang, Kenton Lee, and Kristina Toutanova. 2019.
\newblock {BERT}: Pre-training of deep bidirectional transformers for language
  understanding.
\newblock In \emph{Proceedings of the 2019 Conference of the North {A}merican
  Chapter of the Association for Computational Linguistics: Human Language
  Technologies, Volume 1 (Long and Short Papers)}, pages 4171--4186,
  Minneapolis, USA.

\bibitem[{Fancellu et~al.(2017)Fancellu, Lopez, Webber, and
  He}]{fancellu2017detecting}
Federico Fancellu, Adam Lopez, Bonnie Webber, and Hangfeng He. 2017.
\newblock Detecting negation scope is easy, except when it isn't.
\newblock In \emph{Proceedings of the 15th Conference of the European Chapter
  of the Association for Computational Linguistics: Volume 2, Short Papers},
  pages 58--63.

\bibitem[{Gururangan et~al.(2020)Gururangan, Marasovi{\'c}, Swayamdipta, Lo,
  Beltagy, Downey, and Smith}]{gururangan2020don}
Suchin Gururangan, Ana Marasovi{\'c}, Swabha Swayamdipta, Kyle Lo, Iz~Beltagy,
  Doug Downey, and Noah~A Smith. 2020.
\newblock Don't stop pretraining: Adapt language models to domains and tasks.
\newblock In \emph{Proceedings of the 58th Annual Meeting of the Association
  for Computational Linguistics}, pages 8342--8360.

\bibitem[{Hossain et~al.(2020)Hossain, Kovatchev, Dutta, Kao, Wei, and
  Blanco}]{hossain2020analysis}
Md~Mosharaf Hossain, Venelin Kovatchev, Pranoy Dutta, Tiffany Kao, Elizabeth
  Wei, and Eduardo Blanco. 2020.
\newblock An analysis of natural language inference benchmarks through the lens
  of negation.
\newblock In \emph{Proceedings of the 2020 Conference on Empirical Methods in
  Natural Language Processing (EMNLP)}, pages 9106--9118.

\bibitem[{Jim{\'e}nez-Zafra et~al.(2020)Jim{\'e}nez-Zafra, Morante,
  Mart{\'\i}n-Valdivia, and
  Ure{\~n}a-L{\'o}pez}]{jimenez-zafra-etal-2020-corpora}
Salud~Mar{\'\i}a Jim{\'e}nez-Zafra, Roser Morante, Mar{\'\i}a~Teresa
  Mart{\'\i}n-Valdivia, and L.~Alfonso Ure{\~n}a-L{\'o}pez. 2020.
\newblock \href {https://doi.org/10.1162/coli_a_00371} {Corpora annotated with
  negation: An overview}.
\newblock \emph{Computational Linguistics}, 46(1):1--52.

\bibitem[{Joshi et~al.(2020)Joshi, Chen, Liu, Weld, Zettlemoyer, and
  Levy}]{joshi2020spanbert}
Mandar Joshi, Danqi Chen, Yinhan Liu, Daniel~S Weld, Luke Zettlemoyer, and Omer
  Levy. 2020.
\newblock {SpanBERT}: Improving pre-training by representing and predicting
  spans.
\newblock \emph{Transactions of the Association for Computational Linguistics},
  8:64--77.

\bibitem[{Khandelwal and Sawant(2020)}]{khandelwal2020negbert}
Aditya Khandelwal and Suraj Sawant. 2020.
\newblock {NegBERT}: A transfer learning approach for negation detection and
  scope resolution.
\newblock In \emph{Proceedings of the 12th Language Resources and Evaluation
  Conference}, pages 5739--5748.

\bibitem[{Konstantinova et~al.(2012)Konstantinova, de~Sousa, Cruz, Ma{\~n}a,
  Taboada, and Mitkov}]{konstantinova-etal-2012-review}
Natalia Konstantinova, Sheila~C.M. de~Sousa, Noa~P. Cruz, Manuel~J. Ma{\~n}a,
  Maite Taboada, and Ruslan Mitkov. 2012.
\newblock \href
  {http://www.lrec-conf.org/proceedings/lrec2012/pdf/533_Paper.pdf} {A review
  corpus annotated for negation, speculation and their scope}.
\newblock In \emph{Proceedings of the Eighth International Conference on
  Language Resources and Evaluation ({LREC}'12)}, pages 3190--3195, Istanbul,
  Turkey. European Language Resources Association (ELRA).

\bibitem[{Lazib et~al.(2019)Lazib, Zhao, Qin, and Liu}]{lazib2019negation}
Lydia Lazib, Yanyan Zhao, Bing Qin, and Ting Liu. 2019.
\newblock Negation scope detection with recurrent neural networks models in
  review texts.
\newblock \emph{International Journal of High Performance Computing and
  Networking}, 13(2):211--221.

\bibitem[{Liu et~al.(2019)Liu, Ott, Goyal, Du, Joshi, Chen, Levy, Lewis,
  Zettlemoyer, and Stoyanov}]{liu2019roberta}
Yinhan Liu, Myle Ott, Naman Goyal, Jingfei Du, Mandar Joshi, Danqi Chen, Omer
  Levy, Mike Lewis, Luke Zettlemoyer, and Veselin Stoyanov. 2019.
\newblock {RoBERTa}: A robustly optimized {BERT} pretraining approach.
\newblock \emph{arXiv preprint arXiv:1907.11692}.

\bibitem[{Mehrabi et~al.(2015)Mehrabi, Krishnan, Sohn, Roch, Schmidt,
  Kesterson, Beesley, Dexter, Schmidt, Liu et~al.}]{mehrabi2015deepen}
Saeed Mehrabi, Anand Krishnan, Sunghwan Sohn, Alexandra~M Roch, Heidi Schmidt,
  Joe Kesterson, Chris Beesley, Paul Dexter, C~Max Schmidt, Hongfang Liu,
  et~al. 2015.
\newblock {DEEPEN}: A negation detection system for clinical text incorporating
  dependency relation into {NegEx}.
\newblock \emph{Journal of Biomedical Informatics}, 54:213--219.

\bibitem[{Merity et~al.(2016)Merity, Xiong, Bradbury, and
  Socher}]{merity2016pointer}
Stephen Merity, Caiming Xiong, James Bradbury, and Richard Socher. 2016.
\newblock Pointer sentinel mixture models.
\newblock \emph{arXiv preprint arXiv:1609.07843}.

\bibitem[{Morante(2010)}]{morante2010descriptive}
Roser Morante. 2010.
\newblock Descriptive analysis of negation cues in biomedical texts.
\newblock In \emph{Proceedings of the Seventh International Conference on
  Language Resources and Evaluation (LREC'10)}.

\bibitem[{Morante and Blanco(2012)}]{morante2012sem}
Roser Morante and Eduardo Blanco. 2012.
\newblock {*SEM} 2012 shared task: Resolving the scope and focus of negation.
\newblock In \emph{*SEM 2012: The First Joint Conference on Lexical and
  Computational Semantics--Volume 1: Proceedings of the main conference and the
  shared task, and Volume 2: Proceedings of the Sixth International Workshop on
  Semantic Evaluation (SemEval 2012)}, pages 265--274.

\bibitem[{Morante and Daelemans(2009)}]{morante2009metalearning}
Roser Morante and Walter Daelemans. 2009.
\newblock A metalearning approach to processing the scope of negation.
\newblock In \emph{Proceedings of the Thirteenth Conference on Computational
  Natural Language Learning (CoNLL-2009)}, pages 21--29.

\bibitem[{Ou and Patrick(2015)}]{ou2015automatic}
Ying Ou and Jon Patrick. 2015.
\newblock Automatic negation detection in narrative pathology reports.
\newblock \emph{Artificial Intelligence in Medicine}, 64(1):41--50.

\bibitem[{Peng et~al.(2018)Peng, Wang, Lu, Bagheri, Summers, and
  Lu}]{peng2018negbio}
Yifan Peng, Xiaosong Wang, Le~Lu, Mohammadhadi Bagheri, Ronald Summers, and
  Zhiyong Lu. 2018.
\newblock {NegBio}: a high-performance tool for negation and uncertainty
  detection in radiology reports.
\newblock \emph{AMIA Summits on Translational Science Proceedings}, 2018:188.

\bibitem[{Ribeiro et~al.(2020)Ribeiro, Wu, Guestrin, and
  Singh}]{ribeiro2020beyond}
Marco~Tulio Ribeiro, Tongshuang Wu, Carlos Guestrin, and Sameer Singh. 2020.
\newblock Beyond accuracy: Behavioral testing of {NLP} models with checklist.
\newblock In \emph{Proceedings of the 58th Annual Meeting of the Association
  for Computational Linguistics}, pages 4902--4912.

\bibitem[{Vincze et~al.(2008)Vincze, Szarvas, Farkas, M{\'o}ra, and
  Csirik}]{vincze2008bioscope}
Veronika Vincze, Gy{\"o}rgy Szarvas, Rich{\'a}rd Farkas, Gy{\"o}rgy M{\'o}ra,
  and J{\'a}nos Csirik. 2008.
\newblock The {BioScope} corpus: biomedical texts annotated for uncertainty,
  negation and their scopes.
\newblock \emph{BMC bioinformatics}, 9(11):1--9.

\bibitem[{Yamada et~al.(2020)Yamada, Asai, Shindo, Takeda, and
  Matsumoto}]{yamada2020luke}
Ikuya Yamada, Akari Asai, Hiroyuki Shindo, Hideaki Takeda, and Yuji Matsumoto.
  2020.
\newblock {LUKE}: Deep contextualized entity representations with entity-aware
  self-attention.
\newblock In \emph{Proceedings of the 2020 Conference on Empirical Methods in
  Natural Language Processing (EMNLP)}, pages 6442--6454.

\end{thebibliography}
\bibliographystyle{acl_natbib}

\appendix



\end{document}